\documentclass[conference]{IEEEtran}
\IEEEoverridecommandlockouts
\usepackage{multirow}
\usepackage{subfigure}
\usepackage{cite}
\usepackage{amsmath,amssymb,amsfonts}
\usepackage{algorithmic}
\usepackage{graphicx}
\usepackage{textcomp}
\usepackage{xcolor}
\usepackage{hyperref}
\usepackage{booktabs}
\DeclareMathOperator*{\argmin}{argmin}
\def\BibTeX{{\rm B\kern-.05em{\sc i\kern-.025em b}\kern-.08em
    T\kern-.1667em\lower.7ex\hbox{E}\kern-.125emX}}

\begin{document}

\title{Multi-Sensor Alignment for Weather Simulations}

 \author{
 \IEEEauthorblockN{Samsad Alam, Devyani Lambhate, Aditya Mohan, Vishal Kumar, and Vaibhav Katewa \thanks{All authors are with the Department of Cyber-Physical Systems at the Indian Institute of Science, Bengaluru. Email IDs: samsadalam@iisc.ac.in, devyani.lambhate@fsid-iisc.in, adityamohan@iisc.ac.in, vishal.kumar1@fsid-iisc.in, and vkatewa@iisc.ac.in.}
 \thanks{This work is supported by Sony India via the Faculty Innovation Award.}
 %\IEEEauthorblockA{\textit{Indian Institute of Science, Bengaluru}}
 }}
 
\maketitle

\begin{abstract}
%The 3D perception performance becomes increasingly critical for autonomous driving during adverse weather, where sensors like Camera, LiDAR, and Radar signals suffer from loss of points, pixel intensity change or noise corruption. To achieve robust performance, it is essential to develop both representative adverse weather datasets alongside robust models capable of extracting information out of weather noise and degradation.  When weather simulation methods are de-
Perception tasks for autonomous vehicles need to work satisfactorily in adverse weather conditions. Due to lack of real-world weather datasets, weather simulations are a promising alternative. To ensure simulations closely mirror real-world weather data, it's crucial that they represent the same weather characteristics, including severity and particle positioning, across different sensors. To achieve this, we propose the Reference Dataset Alignment Method (ReDAM) for weather intensity alignment in fog and Unified-weather-edit (inspired by Weather-edit~\cite{weather-edit}) for particle positioning alignment in rain and snow. We validate both alignment methods using statistical and geometrical tests, respectively. We find that 3D detection models for non-aligned versions tend to be overly optimistic as compared to aligned versions. We also show the aligned-multi-sensor simulation's effectiveness for achieving robustness for 3D object detection task by finetuning existing sensor fusion models on it.

%We additionally validate weather alignment in the Simulated dataset and it's effectiveness for three weather types (fog, rain, and snow) by finetuning sensor fusion based 3D object detection models on it.

%These sensor dedicated simulation methods do not guarantee consistency based on weather severity and 3D scene geometry. Our model Unified-weather-edit (Uni-Weather-edit), based on weather-edit, proposes alignment based on location of weather particles in different sensors (geometry alignment) and severity of weather, along with the view and temporal alignment proposed by weather edit. To achieve geometical alignment we generate weather particles in 3-D would coordinate and then project them to individual sensor domains. We show how non aligned datasets tend to give over-optimistic results. Uni-weather-nuscenes, a dataset generated using uni-weather-edit gains 29\% mAP w.r.t. to the model trained on nuscenes. 

\end{abstract}

%\begin{IEEEkeywords}
%Autonomous driving, LiDAR, Camera, Simulation, Weather, Alignment
%\end{IEEEkeywords}

\section{Introduction}
In autonomous driving, 3D perception aims to understand the surrounding environment in full 3D using multiple sensors such as LiDAR and Camera. A key task is 3D object detection, which predicts 3D bounding boxes for objects like vehicles, pedestrians, and cyclists. For reliable deployment, such systems must operate robustly under adverse weather conditions, including fog, rain, and snow. However, widely used datasets contain only a limited number of adverse weather samples, leading to poor generalization in such scenarios.

%such as nuScenes \cite{caesar2020nuScenesmultimodaldatasetautonomous}, KITTI \cite{geiger2013vision}, and Waymo \cite{sun2020scalability} contain only a limited number of adverse weather samples, leading to poor generalization in such scenarios.

To mitigate data scarcity, prior works \cite{improving} employ simulation-based approaches to generate adverse weather data. While effective in increasing representation of adverse weather conditions in the dataset, these methods often simulate each sensor independently, resulting in misaligned representations of weather characteristics such as severity and particle distribution across modalities. This inconsistency deviates from real-world conditions. In this work, we propose sensor-aligned weather simulation methods to ensure alignment in weather severity and particle position. 

 Some of the previous works have proposed weather simulated datasets. Nuscenes-C \cite{dong2023benchmarking} is a commonly used corrupted dataset that uses physics based weather simulation methods for LiDAR and image enhancements from image processing or Computer-Vision models to simulate visually realistic weather Camera Images, but it does not address sensor alignment. MSC-Bench\cite{msc} also introduces weather simulations using a snow simulator \cite{hahner2022lidar} and a fog simulator\cite{hahner2021fog} for LiDAR and then matches weather severity in Camera using snowfall levels and fog parameters. Weather-edit\cite{weather-edit} proposes aligning particle positions in different views and temporal frames of Camera Images, but does not handle cross-sensor alignment. Although some of the approaches focus on weather severity alignment, to the best of our knowledge, there are no existing approaches to address weather particle positioning alignment across sensors.
 %\studies address intensity level alignment but they do not address geometry based alignment.

%~\cite{hahner2021fog}
Each weather type requires a tailored sensor alignment method. Homogeneous weather mediums like fog or dust, where a large number of weather particles are uniformly distributed in the 3D space requires only intensity alignment. In contrast, non-homogeneous mediums like snow or rain, where weather particles are much bigger and sparser, require position alignment of weather particles along with intensity based alignment. For example, a rain streak at $(x,y,z)$ positions in 3D space affects both Camera and LiDAR rays passing through it. To achieve this, we propose {Re}ference {D}ataset {b}ased {A}lignment {M}ethod (ReDAM) for homogeneous weather mediums and Unified-weather-edit (based on Weather-edit\cite{weather-edit}) for non-homogeneous mediums. Our key contributions are listed below.

%Previous works use a dedicated simulation method for each sensor, which does not ensure the representation of same 3D scene in the generated weather simulation. In a real weather scenario, a rain streak at (x,y,z) positions in 3D scene affects both Camera and LiDAR. This phenomenon cannot be reproduced with independent sensor fusion models. Some of the previous works have proposed sensor alignment based on weather severity\cite{dong2023benchmarking}, which only pairs the sensors data based on intensity of the weather, rather than matching the exact number of weather particles or the positions of the weather particles. We have addressed this sensor-alignment problem using a intensity and geometry aware alignment in our Unified-weather-edit model. This model is based on weather-edit\cite{weather-edit}, which generates view and temporal aligned Camera images. To achieve intensity and geometry aware alignment we generate weather particles in 3-D world coordinates and then project them to their individual sensor spaces. We perform this simulation on nuScenes-mini data and generate Uni-weather-nuScenes-mini dataset. We have shown that an mAP gain of upto 29\% can be achieved on simulated weather dataset along with retaining the performance on clean dataset.   \\

\begin{enumerate}
    \item \emph{Re}ference \emph{D}ataset based \emph{A}lignment \emph{M}ethod (ReDAM) which aligns the fog severity in simulated LiDAR and Camera using a real-world weather dataset as reference.   %matches the LiDAR weather intensity by tuning weather parameters with an aim to reduce distance between the real and simulated intensity distributions. For Camera we use real images to induce similar weather effect using style transfer.

    \item Unified-weather-edit model to achieve (a) position alignment of weather particles in simulated LiDAR and Camera. This alignment uses a common 3D world coordinate representation for particles. (b) Temporal alignment of particles in 3D world coordinates, which results in temporal alignment in both sensors. %and then project them to their individual sensor co-ordinates from where the simulation is carried out independently in each sensor. We also bring temporal alignment in LiDAR similar to temporal-view alignment in image as done by~\cite{weather-edit}
    %\item Unified-weather-edit model achieve position and intensity based sensor aligned rain and snow simulations
    %\item Unified-weather-edit model introduces temporal alignment of particles in world coordinates, which results in temporal alignment in both the sensors.
    \item Alignment verification using KS-Statistics \cite{ks_test} and image quality assessment in ReDAM;  Geometric verification of particle alignment in Unified-weather-edit with re-projection error, and perceptual verification with sensor fusion based 3D object detection. 
\end{enumerate}

\section{Weather Simulation and Sensor Alignment}
In this section, we first brief about our weather simulation methodology and then present sensor alignment methods.

\begin{figure*}[htbp]
    \centering
    \begin{minipage}[c]{0.22\textwidth}
        \centering
        \includegraphics[width=\linewidth]{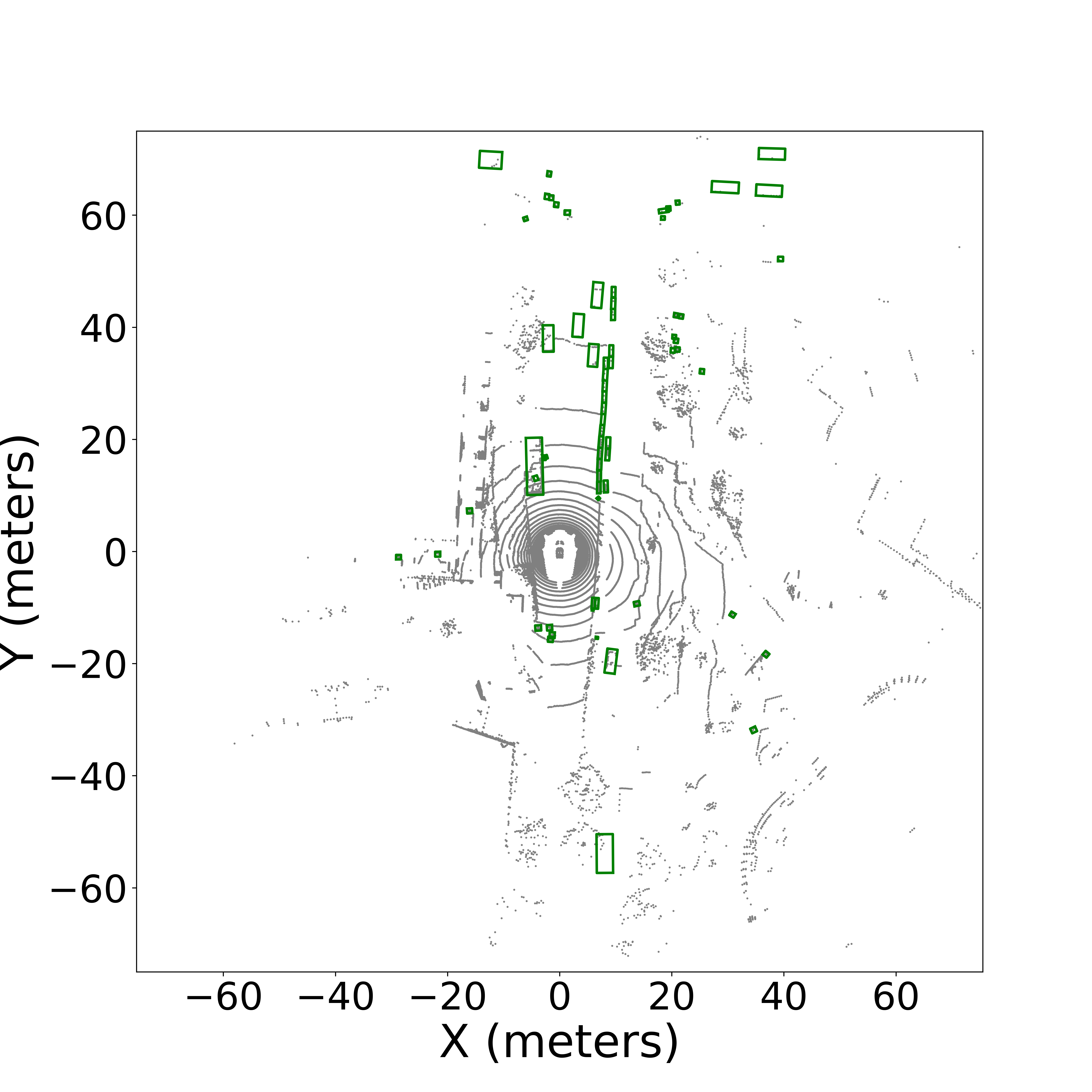}
        \vspace{1mm}
        {\small (a) Clean LiDAR BeV}
    \end{minipage}
    \hfill
    \begin{minipage}[c]{0.22\textwidth}
        \centering
        \includegraphics[width=\linewidth]{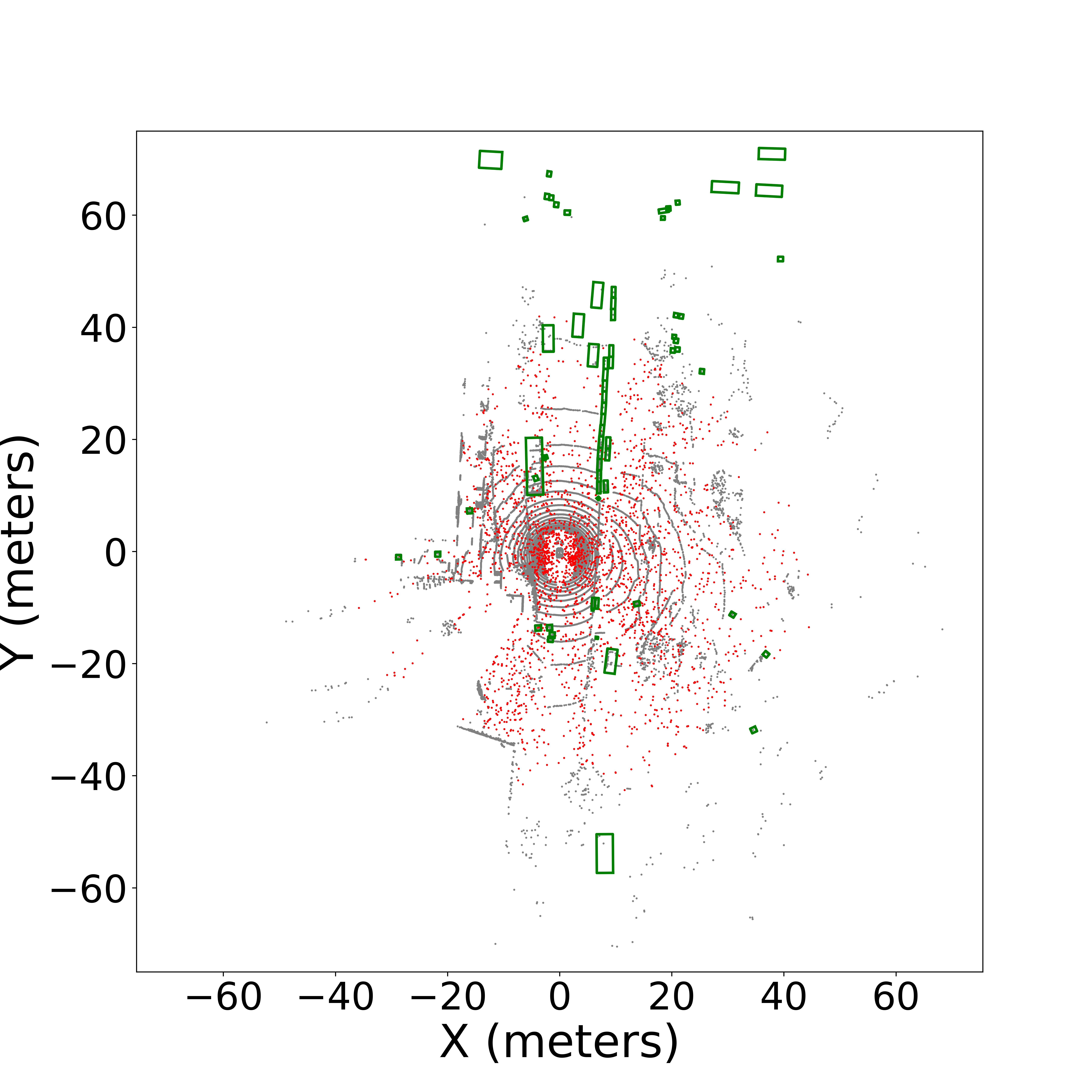}
        \vspace{1mm}
        {\small (b) Simulated LiDAR BeV}
    \end{minipage}
    \hfill
    \begin{minipage}[c]{0.26\textwidth}
        \centering
        
        \vspace*{4mm}
        \includegraphics[width=\linewidth,height=3.1cm]{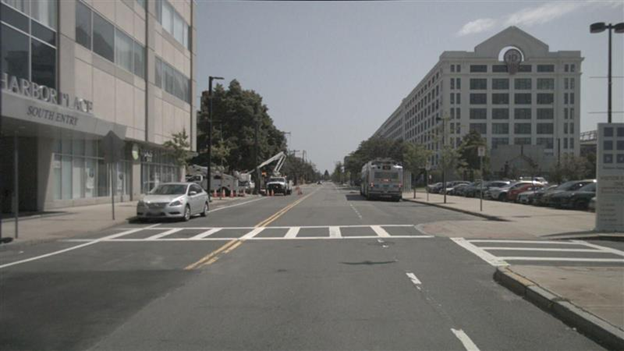}
        \vspace*{0mm}
        
        {\small (c) Clean Image}
    \end{minipage}
    \hfill
    \begin{minipage}[c]{0.26\textwidth}
        \centering
        
        \vspace*{4mm}
        \includegraphics[width=\linewidth,height=3.1cm]{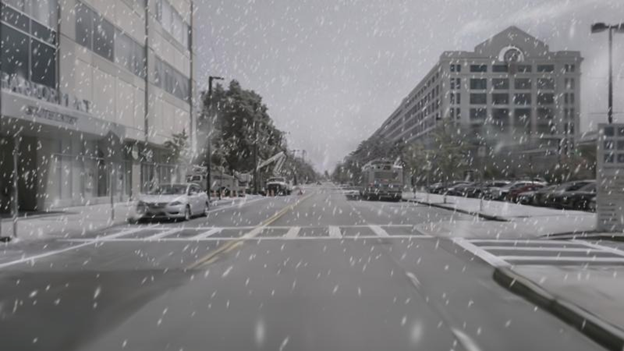}
         \vspace*{0mm}
        
        {\small (d) Simulated Snow Image}
    \end{minipage}
    \caption{Simulation of snowy weather on a LiDAR sample ((a)-(b)) and a Camera image sample ((c)-(d)). Red points in (b) appear due to backscatter from snowflakes. There is also a loss of points far away from LiDAR, noticeable by comparing gray points in (a) and (b). (d) contains both the background edit effect (snow deposit on the footpath) and rendered snowflake particles.}
    \label{fig:lidar_weather_sim}
\end{figure*}

\subsection{Weather Simulation}

% \textcolor{red}{Our objective is to generate weather simulations that are consistent across LiDAR point clouds and camera images. This requires two properties: (i) similar background weather effects such as lighting and visual style, and (ii) consistent spatial locations of weather particles. LiDAR is unaffected by background style as it has its own source of light. It's variation arise solely from the attenuation of laser in the medium and backscatter from weather particles (raindrops, snowflakes, etc.). Camera is impacted by both lighting and background style. To achieve alignment of weather severity and particle position, we discuss here the simulation methods that specifically account for inducing these effect in their respective sensors.}

\subsubsection*{\textbf{LiDAR weather simulation}} 
\label{subsub:LiDAR_weather_simulation}
We use physics-based models for LiDAR simulations as they provide parametric controllability of weather severity. To capture the effect of atmospheric media on LiDAR reflections, we leverage the LiDAR impulse response given by~\cite{2011Rasshofer_automotive_LiDAR} 

\begin{equation}
P_{rec}(R) = C_A \int_{0}^{2R/c} P_{tr}(t) H\left(R - \frac{c t}{2}\right) dt,
\label{eqn:LiDAR_eqn}
\end{equation}
where $P_{rec}(\cdot)$ is the
received power, $R$ is the distance from LiDAR, $P_{tr}(\cdot)$ is the transmitted power, $H(\cdot)$ is the channel response, $C_A$ is the sensor constant and $c$ is the speed of light.
% \begin{align*}
%     P_{tr}(t) = 
%             \begin{cases} 
%             P_0 \sin^2 \left( \frac{\pi}{2\tau_H} t \right) & 0 \le t \le 2\tau_H \\
%             0 & \text{else}
%             \end{cases}
%     \quad P_0 = \text{peak power}, \\  \tau_H = \text{half pulse width}
% \end{align*}

We model the weather effects by selecting appropriate $H (R)$ for fog, rain, and snow. Target object response is always modeled using impulse function. Fog response is modeled in~\cite{hahner2021fogsimulationrealLiDAR} as a homogeneous medium with the step function response characterized by attenuation factor $\alpha$ and fog reflectivity $\beta$. Snow and rain are modeled by ~\cite{hahner2022LiDARsnowfallsimulationrobust} and ~\cite{kilic2021LiDARlightscatteringaugmentation}, respectively as discrete media with particles populating the field of view of LiDAR. Each snow/rain particle is characterized by an occlusion ratio signifying portion of beam that particle blocks and reflectivity of the particle. Each particle's response is characterized by impulse function just like target object. The received power is computed by superposing response from each particle in the beam. Upon computation of received power in a beam we detect where the peak occurs and that becomes our modified location and the peak power becomes modified intensity. \autoref{fig:lidar_weather_sim} (a) and (b) show the snowy weather simulation on a sample point cloud.

\subsubsection*{\textbf{Camera weather simulation}}\label{subsubsec:cam_weather_sim}
To simulate weather in Camera images, we follow methods described in TSIT~\cite{tsit} and Weather-edit.
\

\noindent \textbf{TSIT} is a style-transfer-based image-to-image translation framework to synthesize images under a desired visual condition by translating base images into a target style domain. In this framework, the base image is treated as the \emph{content image}, providing the semantic structure and spatial layout of the scene, while images from the target domain serve as \emph{style references} that capture the corresponding visual characteristics of the desired condition.
%We leverage this style transfer capability to generate images with the same style as the target domain. In our work we use TSIT for fog simulation where the content images come from the set on which simulation is to be done and  target domain are real foggy images.

%The architecture employs a two-stream design consisting of a content stream and a style stream, which extract multi-scale feature representations from the content and style images respectively. These feature representations are subsequently fused within the generator through feature transformation modules to inject the desired appearance attributes while preserving the underlying scene structure. Consequently, the model is able to generate images in the target visual domain while maintaining the semantic consistency and geometric integrity of the original scene. In our case, this framework can be directly used to generate images under desired adverse weather conditions, such as rain, fog, or snow, with intensity levels corresponding to those present in the style images.
%\color{blue} Weather edit has two components(A) background (b) particle \color{black}

\noindent \textbf{Weather-edit} is another model designed to simulate weather effects across multiple views and frames, while maintaining consistency in both multi-view and temporal (multi-frame) dimensions. This is achieved through a two-stage process.
%Since autonomous driving dataset images are captured across multiple views and multiple time steps, processing them individually will cause inconsistent weather effects. Weather-edit operates in two stages to achieve multi-view and multi-time alignment of Camera images. 

(a) The first step is \textbf{background editing}, where a diffusion model is used to generate weather-specific background effects, such as, whitish snowy or cloudy rainy background. 
%It uses a clip encoder that provides a weather-specific prompt as a conditional input to the diffusion model to create semantically aware background effects (e.g. snow on trees, rain on roads). Inside the Diffusion model's UNet, temporal and view attention is used to ensure multi-time and multi-view consistency.

%It finetunes one diffusion model with three LoRA adapters, one for each weather types $i=fog/rain/snow$ for background editing. A clip encoder embeds a text prompt describing the weather background into weather embedding $\mathbf{p}$ that is provided as a conditional input to diffusion model along with semantic map $M$ to create semantically aware background effect (snow on trees, rain on roads). At this step a temporal and view attention is used to ensure multi-frame and multi-view consistency.  \\

(b) The next step is \textbf{particle rendering}, which models the weather particles with 4D Gaussian field. Each weather type is represented by Gaussian particles with attributes $A_i = \{C_i, P_i, R_i, S_i, O_i\}$ which represent color, position, rotation, scale, and opacity of Gaussian $i$, respectively, chosen to mimic the real weather properties of snowflakes or raindrops. Since the weather particles are simulated in a unified 3D world coordinate system and then shared across all views, this inherently ensures multi-view consistency. Further, the motion of the particles in 3D world coordinate is approximated using a constant terminal velocity vector $\mathbf{D}$, which is used to update the position of each particle as: $\mathbf{P}_{world} (t + \Delta t) = \mathbf{P}_{world} (t) + \mathbf{D} \Delta t $. This results in temporal consistency in particles across frames. \autoref{fig:lidar_weather_sim} (c) and (d) show the snow weather simulation on a Camera image sample.

%\autoref{fig:LiDAR_weather_sim} (c),(d) show the weather simulation by Weather-edit on a sample image.

%\color{blue}explain 3D world coordinates\color{black}

%%%% UNCOMMENT/COMMENT TO ADJUST PAGE SIZE   %%%%%%%%%%%%%
%Weather-edit also simulates dynamic weather particles with the help of 4D gaussians. Each weather type is represented by Gaussian particles with attributes $A_i = \{C_i, P_i, R_i, S_i, O_i\}$ where $C_i, P_i, R_i, S_i, O_i$ are color, position, rotation, scale and opacity of gaussian $i$ respectively chosen  to mimic real weather properties of snowflakes or raindrops. 
% These attributes are chosen  to mimic real weather properties of snowflakes or raindrops. To vary severity of weather, we change number of particles and their attributes. 

% Since this method provides control of particle position and their movement, we leverage this to create particle aligned LiDAR-camera simulation.
%%%% UNCOMMENT/COMMENT TO ADJUST PAGE SIZE   %%%%%%%%%%%%%

% \begin{equation}
%     \mathbf{P}_j(t+\Delta t) = \mathbf{P}_j(t) + \mathbf{D}_j \Delta t,
% \end{equation}
% where $\mathbf{P}_j(t) = [x(t), y(t), z(t)]^\top$ and $\mathbf{D}_j = [D_x, D_y, D_z]^\top$. This gives falling particle effect consistent across frames.

\begin{figure*}[htbp]
    \centering
    \begin{minipage}[c]{0.455\textwidth}
        \centering
        \vspace*{10mm}
        \includegraphics[width=\linewidth]{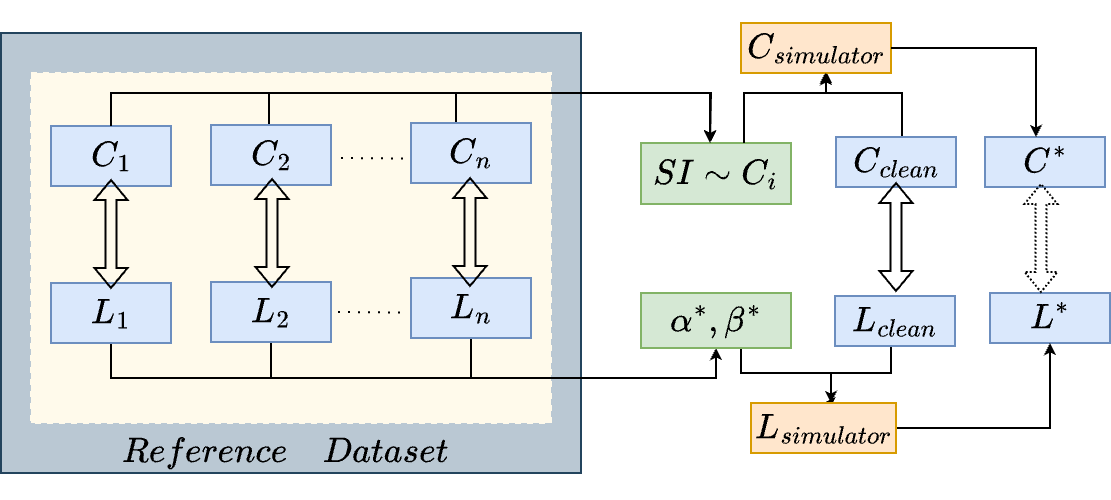}
        
        \vspace*{0mm}
        {\small (1) ReDAM alignment}
    \end{minipage}
    \hfill
    \begin{minipage}[c]{0.525\textwidth}
        \centering
        \includegraphics[width=0.9\linewidth,height=4.5cm]{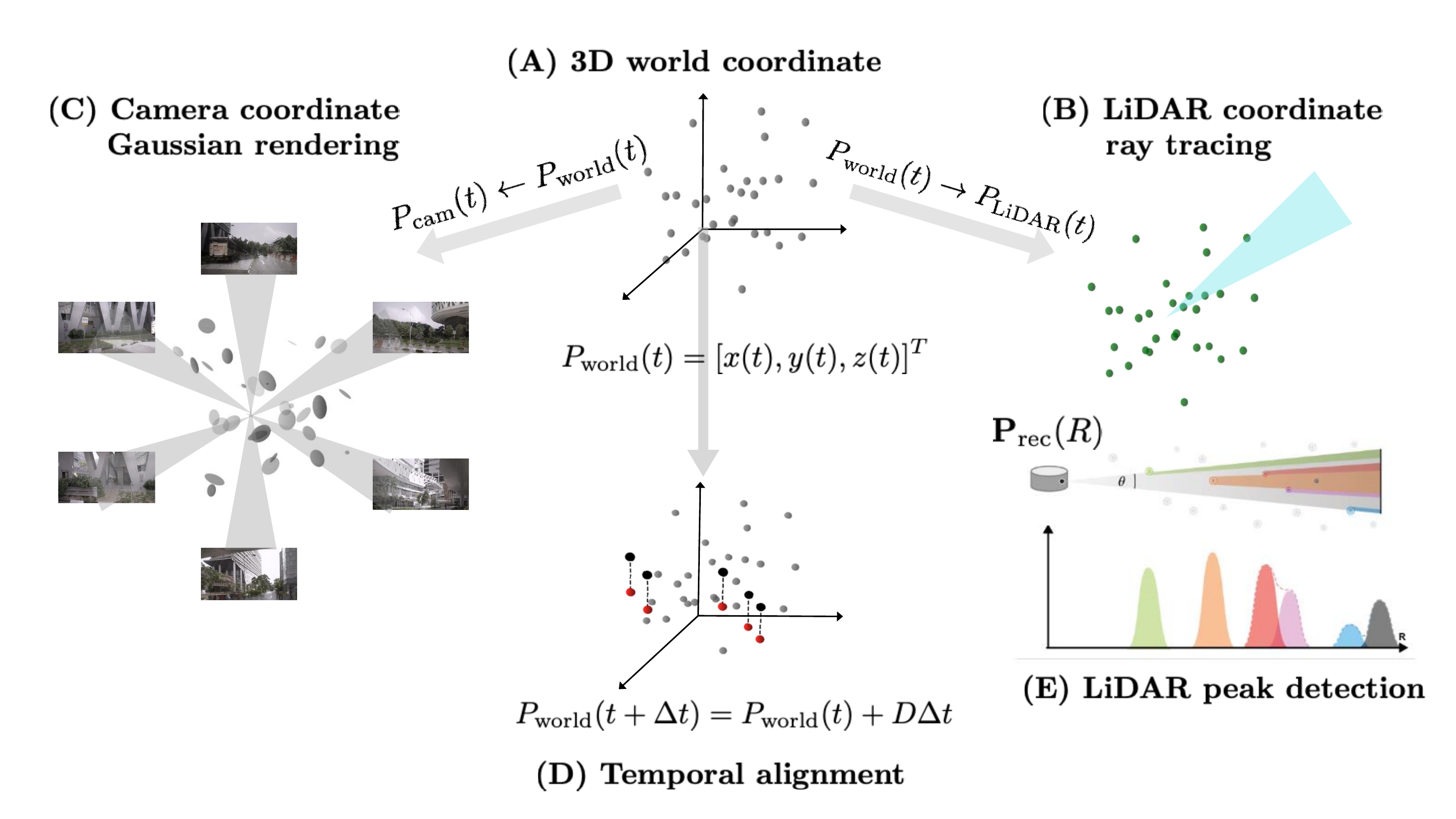}
         \vspace{1mm}
        {\small \phantom{aaa} (2) Unified-weather-edit alignment}
    \end{minipage}
    \caption{(1) \textbf{ReDAM}: $\{C_{1},\cdots,C_{n}\}$ and  $\{L_{1},\cdots,L_{n}\}$ represent Camera and LiDAR data from the reference dataset. $SI$ is a Sample Image from Camera data. LiDAR and Camera simulators $(C_{simulator},L_{simulator})$ take the clean samples $(C_{clean},L_{clean})$ and generate the weather simulated samples $(C^*,L^*)$.
    %$C_{clean}$ and $L_{clean}$ are the clean Camera and LiDAR samples on which we need to perform the simulation. $C_{simulator}$ and $L_simulator$ are the Camera and LiDAR simulators which result in weather simulated Camera and LiDAR samples $C^*$ and $L^*$. 
    (2) \textbf{Unified-weather-edit}: Particle positions in 3D world coordinate (\textbf{A}) are transformed to  LiDAR coordinate (\textbf{B}) and  Camera coordinate (\textbf{C}).
    %with the help of transformation matrices (shown by arrows from (\textbf{A}) to (\textbf{B}) and (\textbf{A}) to (\textbf{C})).
    Transformed coordinates are used for particle simulation in each sensor.} 
    \label{fig:alignment_methods}
\end{figure*}

\subsection{Weather Alignment Across Sensors}
We propose two different methods of alignment between LiDAR and Camera: \textbf{ReDAM for weather severity alignment} that aligns weather severity for homogeneous weather medium like fog, and \textbf{Position-based particle alignment using Unified-weather-edit} that aligns particle positions for non homogeneous medium like snow and rain.

\subsubsection{\textbf{ReDAM for Fog}}
To achieve fog severity alignment between simulated LiDAR and Camera, we propose to leverage a reference dataset that has real and aligned (LiDAR-Camera) fog data. We choose \emph{Seeing Through Fog} (STF)~\cite{Bijelic_2020_STF} dataset for this purpose. ReDAM consists of two steps -
(i) Aligning the simulated LiDAR weather with the STF LiDAR data 
(ii) Aligning the simulated Camera weather with the STF Camera data.

Since the STF dataset contains aligned LiDAR and Camera observations under real fog, aligning our simulations to STF implicitly produces aligned weather simulations in both sensors. Next, we explain the above two steps in detail. \\

% This method uses a reference dataset having real or aligned weather in both LiDAR and camera and is applicable to homogeneous medium only. Here we have used fog which is assumed in literature to be homogeneous medium~\cite{2011Rasshofer_automotive_LiDAR}, ~\cite{hahner2021fog}. We choose \emph{Seeing Through Fog}~\cite{Bijelic_2020_STF} dataset for this purpose which contains fog in both LiDAR (refer to as STF LiDAR) and camera (refer to as STF camera) captured under real weather. The alignment process in LiDAR and camera simulation illustrated in \autoref{fig:alignment_methods} is then done as follows - 

\noindent \textit{Severity Alignment of Simulated LiDAR with STF LiDAR:} %\color{blue} severity alignment of Simulations with STF \color{black}

\noindent Recall that we use the physics-based model \eqref{eqn:LiDAR_eqn} for LiDAR fog simulation. In this model, the fog severity is captured by parameters $\alpha$ and $\beta$. Thus, we propose to tune these parameters to minimize the \textbf{KS-Statistics}~\cite{ks_test} between the intensity distributions of simulated and STF point cloud. We estimate the empirical cumulative distribution function (ECDF) $F_{real} (x)$  of intensity $X$ of STF point cloud and $F_{sim, (\alpha, \beta)} (x)$ of simulated point cloud from respective samples, and compute the parameter values as:
\begin{align*}
    (\alpha^*, \beta^*) = \argmin_{(\alpha, \beta)} \left[ D_{KS} = \sup_{x} | F_{sim, (\alpha, \beta)} (x) - F_{real}(x)| \right]
\end{align*} 
We then use the tuned parameters ${\alpha^*, \beta^*}$ to simulate fog on our desired dataset as mentioned in Section  \ref{subsub:LiDAR_weather_simulation}. 

\noindent \textit{Severity Alignment of Simulated Camera with STF Camera:} 
% Recall that we use TSIT for Camera fog simulation. TSIT uses a target style image that captures the weather conditions and severity that needs to be simulated. For each of the clean image from the nuscenes dataset, we propose to independently sample the target style image randomly from the corresponding STF fog level dataset and pass them through TSIT architecture to generate the simulated images.
% % We propose to pick  images from the STF dataset as target style images and use them to generate simulated images. 
% This ensures that the fog severity of the simulated weather images aligns with the STF data. 
We employ TSIT for fog simulation in camera images, where a target style image encodes the desired weather conditions and severity. For each image from the nuScenes-mini~\cite{caesar2020nuscenesmultimodaldatasetautonomous}dataset, a corresponding target style image is randomly sampled from the STF dataset at the same fog level and passed through the TSIT architecture to generate the simulated image. This procedure ensures that the severity of the fog in the simulated image and the STF dataset is closely aligned.
%to achieve the To ensure that the fog intensity in the simulated Camera images matches that of the STF dataset, we randomly sample corresponding fog-level images from the STF dataset \cite{Bijelic_2020_STF} and used them as style reference images. These were combined with content reference images from the nuScenes \cite{caesar2020nuScenesmultimodaldatasetautonomous} dataset using a style transfer framework~\cite{tsit} to generate foggy images with LiDAR-consistent fog levels.
The overall ReDAM process is illustrated in \autoref{fig:alignment_methods}-(1). To generate simulations with different severities, the ReDAM process can be repeated for real-weather reference subsets representing different weather severities.

% To ensure the same fog intensity level as present in the stf dataset in the camera images, we randomly sampled the corresponding fog level image from the stf dataset and used that as a style reference image to generate the LiDAR consistent fog levels on top of content reference image from nuscenes dataset using style transfer framework~\cite{tsit} to generate aligned foggy images.

%Since the STF dataset contains aligned LiDAR and camera observations under real fog, aligning our simulations to STF implicitly produces aligned weather simulations in both sensors.

\subsubsection{\textbf{Unified-weather-edit for Snow and Rain}}
\label{subsub:uni_weather_edit}
 %We propose -- to align multiple sensors\color{black}
 
 To align particle (snowflake/raindrop) positions between LiDAR and Camera, we propose to use a unified 3D World coordinate framework similar to that used in Weather-edit. Since LiDAR is largely uncorrelated with image background changes~\cite{Weitkamp2005LiDAR}, we adapt only the particle position alignment step described in Subsection \ref{subsubsec:cam_weather_sim}. We first determine the number of particles to be generated based on the snowfall/rainfall rate and the terminal velocity. Then, we render each particle in the 3D World coordinate by randomly sampling its coordinates $[x,y,z]$ uniformly in the field of view of sensors (refer to \autoref{fig:alignment_methods}, 2-A). 
 %giving us $N$ co-ordinate points - $\mathbf{P}_{world} \in \mathbb{R}^{N \times 3}$ , (2)). $N$ is chosen according to the
 Next, we transform these particle positions in LiDAR and Camera coordinates with the following transformation equations
 \begin{align*}
     \mathbf{P}_{s} &= \mathbf{W}_{world \rightarrow s} \mathbf{P}_{world} \\
     %\mathbf{P}_{cam} &= \mathbf{W}_{world \rightarrow cam} \mathbf{P}_{world},
 \end{align*}
 where $s \in [LiDAR, Camera]$, $\mathbf{W}_{world \rightarrow s}$ is $\mathbb{R} ^ {3 \times 3}$ World to LiDAR and Camera transformation matrices, respectively. Since the LiDAR and Camera particle positions are derived from a unified world coordinate, this ensures spatial particle alignment between LiDAR and Camera. Further, the temporal consistency of the world coordinates intrinsically results in temporal alignment in each of Camera and LiDAR. 
These positions are then utilized independently in the respective sensor to simulate particle effects, as explained next.

 \noindent \textit{LiDAR simulation: } We place the particles at location $\mathbf{P}_{LiDAR}$. The diameter of the particles is chosen as described in Subsection \ref{subsub:LiDAR_weather_simulation}. We find particles in each LiDAR beam and compute occlusion ratio for each particle intersecting the beam. Then, we compute the response from each particle as described in Subsection \ref{subsub:LiDAR_weather_simulation} to simulate the weather.

 \noindent \textit{Camera simulation: } We place Gaussian means at the particle locations $\mathbf{P}_{cam}$ as illustrated in \autoref{fig:alignment_methods}, (2-C). Then, we select associated attributes like scale, rotation, opacity, and color. These particle Gaussians are subsequently integrated with scene Gaussians from weather-edit and rendered jointly to generate images containing simulated weather particles.

\section{Experimental setup and Results}
% In ReDAM method to tune parameters $\alpha, \beta$ and to sample style images we have used STF's clean and fog splits. Clean split has 2183 image and lidar samples and fog has 1049 light fog and 881 dense fog samples. Here we have considered left stereo view images only. For particle alignment we have used nuScenes mini dataset as STF has only two stereo views and weather-edit doesn't work reliably with small number of views. We have used mini subset instead of full trainval set to avoid huge computational cost involved in weather simulation and weather-edit 3D reconstruction pipelines. nuScenes-mini has all sensor suits as trainval but contains only subset of scenes - 10 scenes with 31,206 sensor frames out of which 404 samples are annotated leading to 18,538 object annotations across 23 object categories. Evaluation of sensor fusion models are also done on nuScenes mini dataset. 

For the ReDAM method, parameter tuning of $\alpha$ and $\beta$, and style image sampling are done using the STF dataset, comprising 2183 clean samples and fog splits with 1049 light and 881 dense fog samples. For particle alignment, we use the nuScenes-mini dataset, as STF’s limited views are insufficient for reliable 3D reconstruction in Weather-edit. To reduce the computational cost of weather simulation and 3D reconstruction in Weather-edit, we adopt the nuScenes-mini subset instead of full trainval set. The nuScenes-mini subset retains full sensor configuration but reduced dataset size (31,206 frames, 404 annotated samples, and 18,538 object annotations from 10 scenes across 23 classes). All sensor fusion evaluations are also conducted on nuScenes-mini.

\subsection*{\bf Simulation parameters} The parameters $\alpha, \beta$ are tuned with grid search over range $\alpha \in \mathcal{A} = \{ 0.001, 0.005, 0.01, 0.02, 0.04, 0.06, 0.08, 0.1\}$ and $\beta \in  \mathcal{B} = 0.023 \times \{ 1.0, 1.2, 1.4, 1.6, 1.8, 2.0 \} / MOR$ where $MOR = 3/\alpha$ defines visibility range. For particle alignment in snow, we have chosen snowfall rate $2.0 mm/hr$ and terminal velocity $\mathbf{D} = 1.5 \hat{z}  \: \: m/sec$ roughly translating to $6 \times 10^6$ particles. For rainfall simulation, we have used rainfall rate $50.0 \: mm/hr$ with terminal velocity $2.0 \: m/sec$ roughly giving $4 \times 10^6$ particles. For relevant calculations, we refer to ~\cite{hahner2022LiDARsnowfallsimulationrobust}. The remaining parameters are same as those used by the respective papers.
\subsection*{\bf Validation of Sensor Alignment}
\subsubsection{Weather Severity Alignment Verification} 
\begin{figure}[htbp]
    \centering
    \begin{minipage}[t]{0.24\textwidth}
        \centering
        \includegraphics[width=\linewidth]{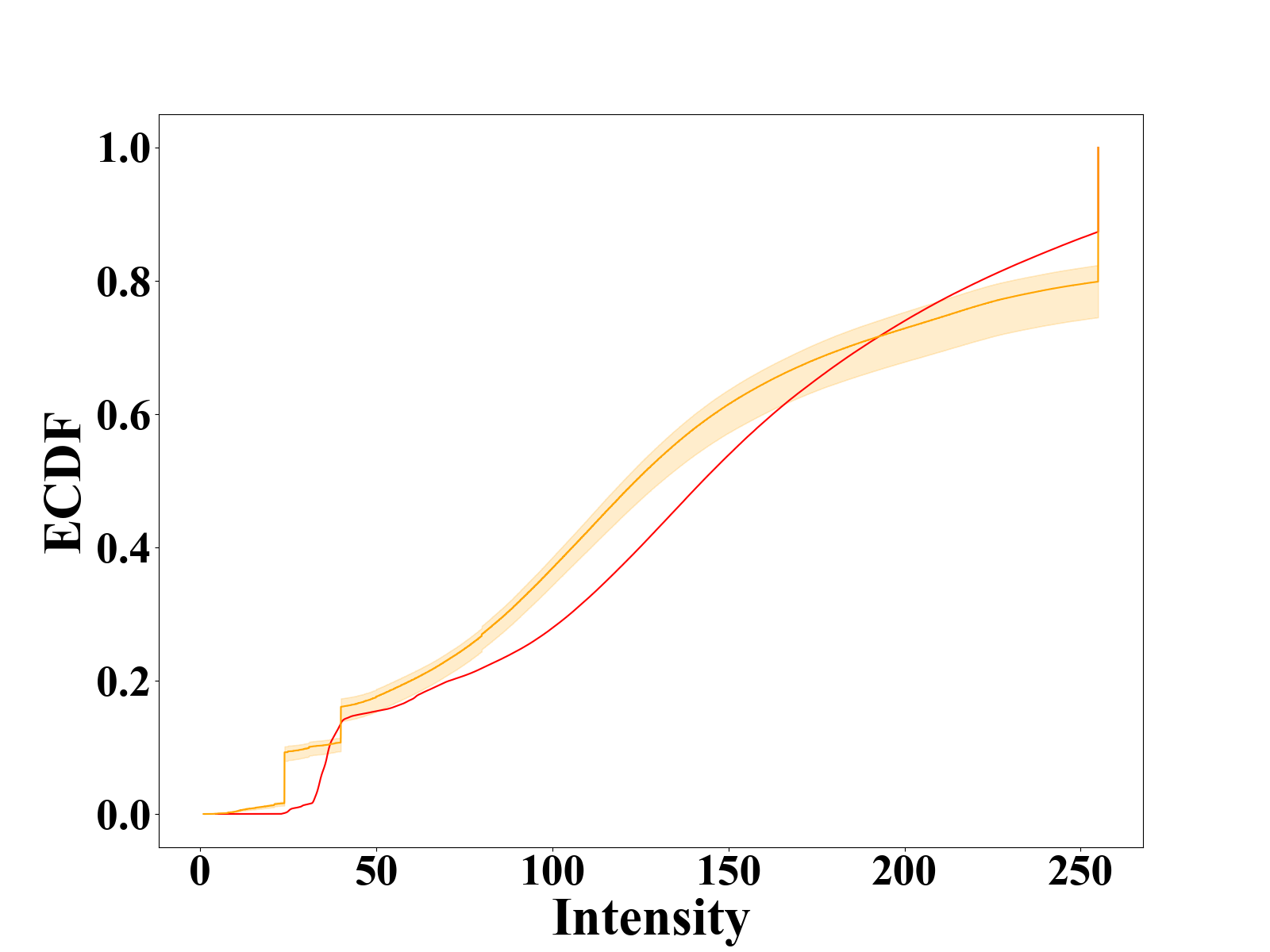}
        % \vspace{1mm}
        {\small (a) Light fog, $D_{KS} = \mathbf{0.092}$}
    \end{minipage}
    %\hfill
    \begin{minipage}[t]{0.24\textwidth}
        \centering
        \includegraphics[width=\linewidth]{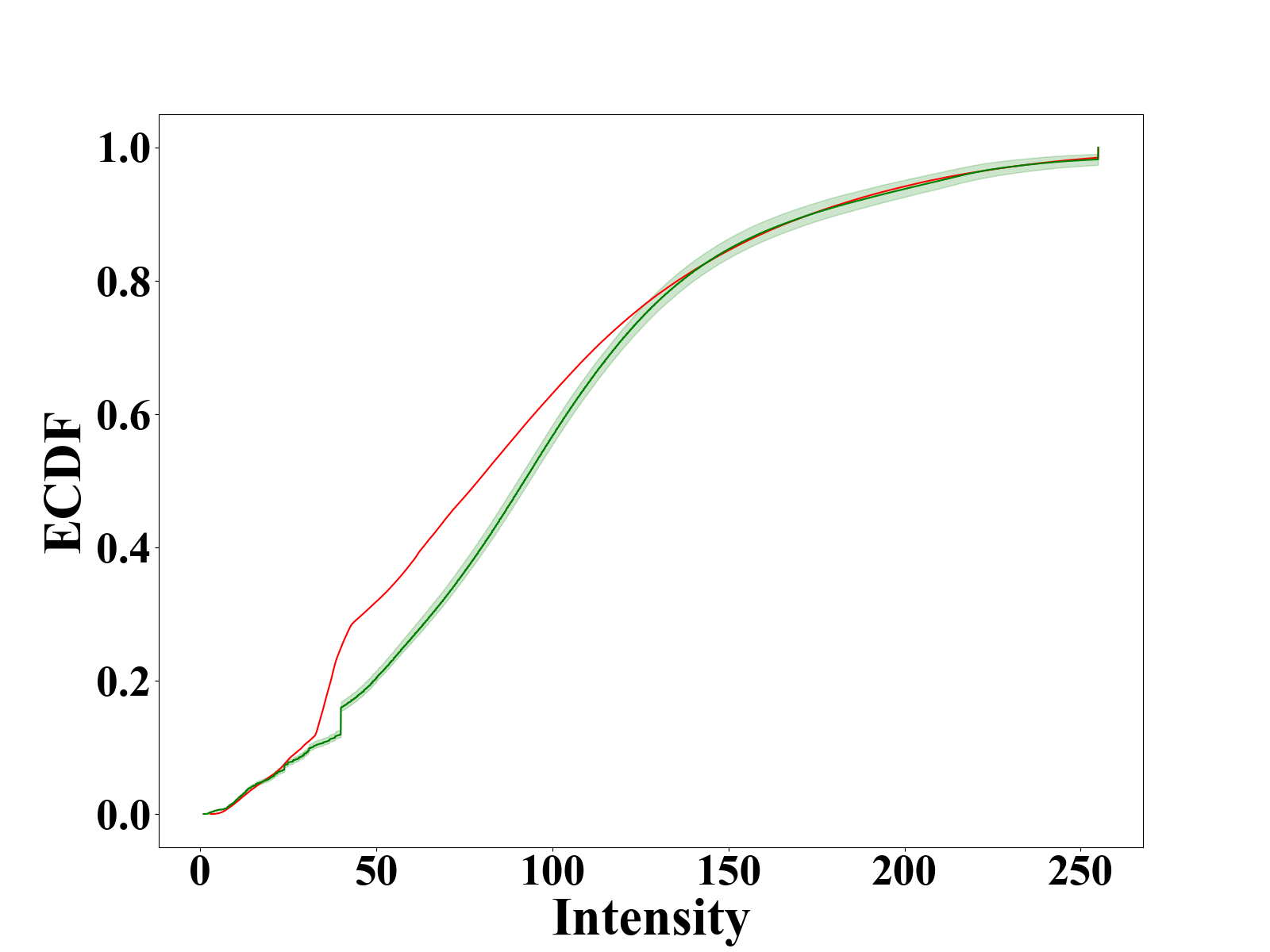}
       % \vspace{1mm}
        {\small (b) Dense fog, $D_{KS} = \mathbf{0.131}$ }
    \end{minipage}
    \caption{Comparison of ECDF plots of simulated and real weather. \textcolor{green}{green} and \textcolor{orange}{orange} bands show intensity ECDF in real weather and \textcolor{red}{red} shows tuned simulated weather ECDF.}
    \label{fig:stats_verification}
\end{figure}

For  ReDAM based fog alignment we obtain $\{\alpha^* = 0.01, \beta^* = 0.023 \times 1.6/MOR = 1.23 \times 10^{-4}\}$ for light fog, and $\{\alpha^* = 0.04, \beta^* = 0.023 \times 2.0/MOR = 6.13 \times 10^{-4}\}$ for dense fog. The parameters are evaluated over all samples in the corresponding split mentioned in STF dataset. The corresponding KS-Statistics values were obtained as $D_{KS} = \mathbf{0.092}$ and $D_{KS} =\mathbf{0.131}$ for light and dense fog, respectively, with the ECDF plots shown in \autoref{fig:stats_verification}. Except at the jump locations, the ECDF shows a close agreement between the simulated fog with real fog in both light and dense conditions. 
For images, we use the  Fr\'echet Inception Distance (FID) \cite{fid} \& Kernel Inception Distance (KID) \cite{kid} metrics between simulated nuScenes foggy images and real STF images with the same fog level. We obtain the FID values of $\mathbf{4.2998}$ \& $\mathbf{3.2772}$ and KID values of $\mathbf{0.1652}$ \& $\mathbf{0.2091}$ for dense and light fog, respectively. The low value of these scores indicates that the generated images match closely with real images. 

\subsubsection{Particle Position Alignment Verification} To verify that weather particles occur at same location, we project these particles in LiDAR onto images and compute the error between projected position and corresponding rasterized position of image Gaussian. We find $\mathcal{N}_{comm}$ as the set of weather particles which remain visible (un-occluded) in both LiDAR and Camera. Then, we compute the re-projection error as  
\begin{align*}
    E = \sum_{i=1}^{|\mathcal{N}_{comm}|} \sum_{j=1}^{N_V} \| \mathbf{u}_{ij} - \hat{\mathbf{u}}(\mathbf{P}_j, \mathbf{X}_i) \|_2^2,
\end{align*}
where $N_V$ is the number of views of the camera, $\mathbf{X}_i$ is the $ith$ lidar point, $\hat{\mathbf{u}}(\mathbf{P}_j, \mathbf{X}_i)$ is projection of $\mathbf{X}_i$ on $jth$ view with projection matrix $\mathbf{P}_j$, and $\mathbf{u}_{ij}$ is the pixel coordinate of the rasterized location of the corresponding Gaussian. The low re-projection error values ($ < 1$ px) shown in \autoref{tab:mmd_weather} validate the correctness of our position alignment.
%\begin{table}[h]
%\centering
%\caption{Re-projection error}
%\begin{tabular}{|lcc|}
%\hline
%Weather & Mean Error (px) & Max Error (px) \\
%\hline
%Snow & $0.7632 $ & $1.3843$ \\
%Rain & $0.7641$ & $1.411$ \\
%%\hline
%\end{tabular}
%\label{tab:reprojection_error}
%\end{table}

\begin{table}[h]
\centering
\caption{Re-projection error and MMD distance}
\label{tab:mmd_weather}
\begin{tabular}{|lccc|}
\hline

{Weather} & {Fog} & {Snow} & {Rain} \\
\hline
Re-projection error & & & \\
\hline
{Mean error}    &  & 0.7632 & 0.7641 \\
{Max error }  &  & 1.3843 & 1.411 \\
\hline 
MMD Distance & & &  \\

\hline
{Aligned}    & \textbf{0.2837} & \textbf{0.3065} & 0.1870 \\
{Unaligned}  & 0.3491 & 0.3607 & 0.1870 \\
\hline

\hline 
\end{tabular}
\end{table}

%\subsubsection{Perceptual Verification with Sensor Fusion}

\subsubsection{Verification using fused features} To verify that aligned simulations are better than unaligned, we compare the MMD~\cite{gretton2012kernel} distance of the fused LiDAR and Camera features between the real and simulated weather samples. We get Camera-LiDAR fused features by auxiliary weather classification model that uses concatenated LiDAR and Camera features and is trained to identify weather type. The lower value of MMD distance between the aligned and real shows a strong correspondence between them.

\subsubsection{Verification using Perception task}
%Downstream tasks offer a practical measure of simulation realism by evaluating model performance on real-world data. 
We assess our alignment methods on the 3D detection task using three SOTA fusion models: BEVFusion (BEV-F) \cite{bevfusion}, Deep Interaction (DI) \cite{deepinteraction}, and Cross-Modal Transformer (CMT) \cite{yan2023cross}.

In Tables \ref{tab:performance_ua_a} and \ref{tab:performance_finetuning}, nuScenes-mini refers to the original clean version of data, nuScenes-mini-snow  and nuScenes-mini-rain refers to the snow and rain simulated data on nuScenes-mini using Unified-weather-edit. nuScenes-mini-fog refers to fog simulation using ReDAM. Pretrained denotes the model trained on nuScenes clear-weather, Finetuned and Finetuned-mix refers to models finetuned on full simulated samples and a mix of 30\% simulated and 70\% clean samples.
% In unaligned versions of the data, the weather particle positions between LiDAR and Camera are not aligned. 

%We have first compared the performance of aligned v/s unaligned version of Unified-weather-edit dataset. In aligned version, the weather particle positions between LiDAR and Camera are aligned. 

We first compare the impact of alignment on 3-D detection tasks as shown in \autoref{tab:performance_ua_a}. The unaligned versions consistently perform better than the aligned version by around $2-3\%$ mAP for rain and around $1\%$ mAP for snow. This can be attributed to the fact that in the aligned dataset, at the positions corresponding to weather particles, information from both sensors is impacted. In contrast, in the unaligned version, partial information from at least one sensor is present.

\begin{table}[htbp]
\centering
\caption{Performance of sensor fusion models on snow and rain data generated from Unified-weather-edit}
\label{tab:performance_ua_a}
% \vspace{1mm}
\setlength{\tabcolsep}{8pt}
\renewcommand{\arraystretch}{1.2}
\begin{tabular}{|l|cc|cc|}
\hline
\multirow{2}{*}{nuScenes-mini-snow} & \multicolumn{2}{c|}{Unaligned} & \multicolumn{2}{c|}{Aligned} \\
\cline{2-5}
 & mAP & NDS & mAP & NDS \\
\hline
BEV-F & 0.4034 & 0.4828 & 0.3943 & 0.4752 \\ 
DI    & 0.4139 & 0.4626 & 0.4010 & 0.4541 \\ 
CMT   & 0.4007 & 0.4800 & 0.3965 & 0.4760 \\
\hline

\multirow{2}{*}{nuScenes-mini-rain} & \multicolumn{2}{c|}{Unaligned} & \multicolumn{2}{c|}{Aligned} \\
\cline{2-5}
 & mAP & NDS & mAP & NDS \\
\hline
BEV-F & 0.4933 & 0.4384 & 0.4649 & 0.5192 \\ 
DI    & 0.5206 & 0.5294 & 0.4926 & 0.5133 \\ 
CMT   & 0.4958 & 0.5316 & 0.4766 & 0.5207 \\
\hline
\end{tabular}
\end{table}
 Next, we show that our generated datasets can be used to achieve robustness to adverse weather. An ideal robust model would improve the performance on weather data (snow, rain, or fog) and retain the performance on the original clean data. As shown by the mAP and NDS scores in Table \ref{tab:performance_finetuning}, the pretrained BeV-Fusion performs very poorly on snowy and rainy data as compared to the clean data.
 
 Next, we finetune the models on our aligned weather dataset and gain $\mathbf{0.13, 0.03, 0.07}$ mAP values, for snow, rain and fog, respectively. Finetuning the models only on simulated-weather datasets results in considerable drop in performance in clear datasets. Therefore, we finetune our models on a mix of weather data and clean data. It is evident from the results that the  mix finetuning performs the best. 

% \begin{table}[htbp]
% \centering
% \begin{tabular}{|l|cc|}
% \hline
% \multicolumn{3}{|c|}{BeV-Fusion-snow}  \\
% \cline{1-3}
%  Mode & nuScenes-mini-clean & nuScenes-mini-snow\\
% \hline
% % Add your data rows here, e.g.:
% Pretrained  & 0.5758, 0.5835 & 0.3943,  0.4752\\
% Finetuned-snowy &  0.5394, 0.5618 & 0.5161, 0.5471 \\
% Finetuned-snowy-mix &  0.5606, 0.5756 & 0.5103, 0.5472 \\

% \hline

% \multicolumn{3}{|c|}{BeV-Fusion-rain}  \\
% \cline{1-3}
%  Mode & nuScenes-mini-clean & nuScenes-mini-rainy \\
% \hline
% % Add your data rows here, e.g.:
% Pretrained  &  0.5758, 0.5835 & 0.4646, 0.5192\\
% Finetuned-rainy & 0.5468, 0.5685 & 0.4909, 0.5287\\
% Finetuned-rainy-mix &  0.5590, 0.5765 & 0.4710, 0.5201
%  \\

% \hline

% \hline
 
% \multicolumn{3}{|c|}{BeV-Fusion-fog}  \\
% \cline{1-3}
%  & nuScenes-mini-clean & nuScenes-mini-fog \\
% \hline
% % Add your data rows here, e.g.:
% Pretrained  &  0.5758, 0.5835 & 0.48,0.52\\
% Finetuned-fog & 0.44, 0.50 & 0.4953, 0.5376\\

% \hline

% \end{tabular}
% \caption{mAP, NDS for pretrained and finetuned versions on snow, rain and fog simulations }
% \label{tab:performance}
% \end{table}

\begin{table}[htbp]
\centering
\caption{MAP and NDS of BEV Fusion on snow, rain and fog simulations}
\label{tab:performance_finetuning}
\setlength{\tabcolsep}{8pt}
\renewcommand{\arraystretch}{1.2}
\begin{tabular}{|l|cc|cc|}
\hline
%\multicolumn{5}{|c|}{Snow}  \\
\cline{1-5}
\multirow{2}{*}{Snow} & \multicolumn{2}{c|}{nuScenes-mini} & \multicolumn{2}{c|}{nuScenes-mini-snow}\\
\cline{2-5}
 & MAP & NDS & MAP & NDS \\
\hline
Pretrained            & 0.5758 & 0.5835 & 0.3943 & 0.4752 \\
Finetuned-snow      & 0.5394 & 0.5618 & \textbf{0.5161} $\uparrow$ & 0.5471 \\
Finetuned-snow-mix  & 0.5606 & 0.5756 & 0.5103 & \textbf{0.5472} $\uparrow$ \\
\hline

%\multicolumn{5}{|c|}{Rain}  \\
\cline{1-5}
\multirow{2}{*}{Rain} & \multicolumn{2}{c|}{nuScenes-mini} & \multicolumn{2}{c|}{nuScenes-mini-rain} \\
\cline{2-5}
 & MAP & NDS & MAP & NDS \\
\hline
Pretrained            & 0.5758 & 0.5835 & 0.4646 & 0.5192 \\
Finetuned-rain      & 0.5468 & 0.5685 & \textbf{0.4909} $\uparrow$ & \textbf{0.5287} $\uparrow$ \\
Finetuned-rain-mix  & 0.5590 & 0.5765 & 0.4710 & 0.5201 \\
\hline

%\multicolumn{5}{|c|}{Fog}  \\
\cline{1-5}
\multirow{2}{*}{Fog} & \multicolumn{2}{c|}{nuScenes-mini} & \multicolumn{2}{c|}{nuScenes-mini-fog} \\
\cline{2-5}
 & MAP & NDS & MAP & NDS \\
\hline
Pretrained      & 0.5758 & 0.5835 & 0.4314   & 0.4991   \\
Finetuned-fog   & 0.4448   & 0.5129   & \textbf{0.5018} $\uparrow$ & \textbf{0.5407} $\uparrow$ \\
Finetuned-fog-mix   & 0.5515   & 0.5725  & 0.4861 & 0.5301 \\
\hline
\end{tabular}
\end{table}

\section{Conclusion and Future Work}

%This work focused on the generating position and intensity aligned weather simulations for 3-D perception tasks. 
We propose ReDAM and Unified-weather-edit to simulate aligned Camera and LiDAR simulations under adverse weather. These alignment methods output more realistic simulations, which can be used by fusion models to achieve robustness via finetuning on the simulations. Unlike the overly optimistic results observed with unaligned data, our simulations present a more grounded assessment.

%.Previous works overlook this weather alignment problem, as they rely sensor dedicated simulation methods. In Unified-weather-edit, we build on top of Weather-Edit to get multi-view, temporal and sensor aligned simulations. We have empirically shown that models trained with unaligned versions tend to be overoptimistic in their performance. We also show the utility of our generated simulated datasets for finetuning BeV Fusion (a 3-D detection model) 

ReDAM and Unified-weather-edit are capable to generate aligned simulations at different weather severity levels. In the future, we plan to provide extensive simulations at different levels for larger dataset (nuScenes trainval). Further, the proposed alignment strategy is easily generalizable to other datasets, sensing modalities (like radar) and other faults (like mixed weather scenarios) which we intend to integrate in future studies.

%\section*{Acknowledgment}
%This work is supported by Sony India via the Faculty Innovation Award.
%We would like to thank Sony R\&D Brussels Lab for funding this project. We would specially like to thank Karin Cvetko Vah for regular meetings, coordination and organizational support through out the course of this project. We would also like to thank the Ministry of Education, Government of India for the graduate scholarship support to S. Alam and A. Mohan.

% Insert References
\bibliographystyle{IEEEtran}
%\IEEEtriggeratref{5}
\bibliography{references}

% Generated by IEEEtran.bst, version: 1.14 (2015/08/26)
\begin{thebibliography}{10}
\providecommand{\url}[1]{#1}
\csname url@samestyle\endcsname
\providecommand{\newblock}{\relax}
\providecommand{\bibinfo}[2]{#2}
\providecommand{\BIBentrySTDinterwordspacing}{\spaceskip=0pt\relax}
\providecommand{\BIBentryALTinterwordstretchfactor}{4}
\providecommand{\BIBentryALTinterwordspacing}{\spaceskip=\fontdimen2\font plus
\BIBentryALTinterwordstretchfactor\fontdimen3\font minus \fontdimen4\font\relax}
\providecommand{\BIBforeignlanguage}[2]{{%
\expandafter\ifx\csname l@#1\endcsname\relax
\typeout{** WARNING: IEEEtran.bst: No hyphenation pattern has been}%
\typeout{** loaded for the language `#1'. Using the pattern for}%
\typeout{** the default language instead.}%
\else
\language=\csname l@#1\endcsname
\fi
#2}}
\providecommand{\BIBdecl}{\relax}
\BIBdecl

\bibitem{weather-edit}
\BIBentryALTinterwordspacing
C.~Qian, W.~Li, Y.~Guo, and G.~Markkula, ``Weatheredit: Controllable weather editing with 4d gaussian field,'' 2025. [Online]. Available: \url{https://arxiv.org/abs/2505.20471}
\BIBentrySTDinterwordspacing

\bibitem{improving}
Y.~Huang, K.~Yu, Q.~Guo, F.~Juefei-Xu, X.~Jia, T.~Li, G.~Pu, and Y.~Liu, ``Improving robustness of lidar-camera fusion model against weather corruption from fusion strategy perspective,'' \emph{arXiv preprint arXiv:2402.02738}, 2024.

\bibitem{dong2023benchmarking}
Y.~Dong, C.~Kang, J.~Zhang, Z.~Zhu, Y.~Wang, X.~Yang, H.~Su, X.~Wei, and J.~Zhu, ``Benchmarking robustness of 3d object detection to common corruptions,'' in \emph{Proceedings of the IEEE/CVF Conference on Computer Vision and Pattern Recognition}, 2023, pp. 1022--1032.

\bibitem{msc}
X.~Hao, G.~Liu, Y.~Zhao, Y.~Ji, M.~Wei, H.~Zhao, L.~Kong, R.~Yin, and Y.~Liu, ``Msc-bench: Benchmarking and analyzing multi-sensor corruption for driving perception,'' \emph{arXiv preprint arXiv:2501.01037}, 2025.

\bibitem{hahner2022lidar}
M.~Hahner, C.~Sakaridis, M.~Bijelic, F.~Heide, F.~Yu, D.~Dai, and L.~Van~Gool, ``Lidar snowfall simulation for robust 3d object detection,'' in \emph{Proceedings of the IEEE/CVF conference on computer vision and pattern recognition}, 2022, pp. 16\,364--16\,374.

\bibitem{hahner2021fog}
M.~Hahner, C.~Sakaridis, D.~Dai, and L.~Van~Gool, ``Fog simulation on real lidar point clouds for 3d object detection in adverse weather,'' in \emph{Proceedings of the IEEE/CVF international conference on computer vision}, 2021, pp. 15\,283--15\,292.

\bibitem{ks_test}
\BIBentryALTinterwordspacing
V.~W. Berger and Y.~Zhou, \emph{Kolmogorov–Smirnov Test: Overview}.\hskip 1em plus 0.5em minus 0.4em\relax John Wiley \& Sons, Ltd, 2014. [Online]. Available: \url{https://onlinelibrary.wiley.com/doi/abs/10.1002/9781118445112.stat06558}
\BIBentrySTDinterwordspacing

\bibitem{2011Rasshofer_automotive_LiDAR}
R.~H. {Rasshofer}, M.~{Spies}, and H.~{Spies}, ``{Influences of weather phenomena on automotive laser radar systems},'' \emph{Advances in Radio Science}, vol.~9, pp. 49--60, Jul. 2011.

\bibitem{hahner2021fogsimulationrealLiDAR}
\BIBentryALTinterwordspacing
M.~Hahner, C.~Sakaridis, D.~Dai, and L.~V. Gool, ``Fog simulation on real lidar point clouds for 3d object detection in adverse weather,'' 2021. [Online]. Available: \url{https://arxiv.org/abs/2108.05249}
\BIBentrySTDinterwordspacing

\bibitem{hahner2022LiDARsnowfallsimulationrobust}
\BIBentryALTinterwordspacing
M.~Hahner, C.~Sakaridis, M.~Bijelic, F.~Heide, F.~Yu, D.~Dai, and L.~V. Gool, ``Lidar snowfall simulation for robust 3d object detection,'' 2022. [Online]. Available: \url{https://arxiv.org/abs/2203.15118}
\BIBentrySTDinterwordspacing

\bibitem{kilic2021LiDARlightscatteringaugmentation}
\BIBentryALTinterwordspacing
V.~Kilic, D.~Hegde, V.~Sindagi, A.~B. Cooper, M.~A. Foster, and V.~M. Patel, ``Lidar light scattering augmentation (lisa): Physics-based simulation of adverse weather conditions for 3d object detection,'' 2021. [Online]. Available: \url{https://arxiv.org/abs/2107.07004}
\BIBentrySTDinterwordspacing

\bibitem{tsit}
L.~Jiang, C.~Zhang, M.~Huang, C.~Liu, J.~Shi, and C.~C. Loy, ``Tsit: A simple and versatile framework for image-to-image translation,'' in \emph{European conference on computer vision}.\hskip 1em plus 0.5em minus 0.4em\relax Springer, 2020, pp. 206--222.

\bibitem{Bijelic_2020_STF}
M.~Bijelic, T.~Gruber, F.~Mannan, F.~Kraus, W.~Ritter, K.~Dietmayer, and F.~Heide, ``Seeing through fog without seeing fog: Deep multimodal sensor fusion in unseen adverse weather,'' in \emph{The IEEE Conference on Computer Vision and Pattern Recognition (CVPR)}, June 2020.

\bibitem{caesar2020nuscenesmultimodaldatasetautonomous}
\BIBentryALTinterwordspacing
H.~Caesar, V.~Bankiti, A.~H. Lang, S.~Vora, V.~E. Liong, Q.~Xu, A.~Krishnan, Y.~Pan, G.~Baldan, and O.~Beijbom, ``nuscenes: A multimodal dataset for autonomous driving,'' 2020. [Online]. Available: \url{https://arxiv.org/abs/1903.11027}
\BIBentrySTDinterwordspacing

\bibitem{Weitkamp2005LiDAR}
C.~Weitkamp, Ed., \emph{Lidar: Range-Resolved Optical Remote Sensing of the Atmosphere}, 1st~ed., ser. Springer Series in Optical Sciences.\hskip 1em plus 0.5em minus 0.4em\relax New York, NY: Springer, 2005.

\bibitem{fid}
\BIBentryALTinterwordspacing
M.~Heusel, H.~Ramsauer, T.~Unterthiner, B.~Nessler, and S.~Hochreiter, ``Gans trained by a two time-scale update rule converge to a local nash equilibrium,'' 2018. [Online]. Available: \url{https://arxiv.org/abs/1706.08500}
\BIBentrySTDinterwordspacing

\bibitem{kid}
\BIBentryALTinterwordspacing
M.~Bińkowski, D.~J. Sutherland, M.~Arbel, and A.~Gretton, ``Demystifying mmd gans,'' 2021. [Online]. Available: \url{https://arxiv.org/abs/1801.01401}
\BIBentrySTDinterwordspacing

\bibitem{gretton2012kernel}
A.~Gretton, K.~M. Borgwardt, M.~J. Rasch, B.~Sch{\"o}lkopf, and A.~Smola, ``A kernel two-sample test,'' \emph{The journal of machine learning research}, vol.~13, no.~1, pp. 723--773, 2012.

\bibitem{bevfusion}
T.~Liang, H.~Xie, K.~Yu, Z.~Xia, Z.~Lin, Y.~Wang, T.~Tang, B.~Wang, and Z.~Tang, ``Bevfusion: A simple and robust lidar-camera fusion framework,'' \emph{Advances in Neural Information Processing Systems}, vol.~35, pp. 10\,421--10\,434, 2022.

\bibitem{deepinteraction}
Z.~Yang, J.~Chen, Z.~Miao, W.~Li, X.~Zhu, and L.~Zhang, ``Deepinteraction: 3d object detection via modality interaction,'' \emph{Advances in Neural Information Processing Systems}, vol.~35, pp. 1992--2005, 2022.

\bibitem{yan2023cross}
J.~Yan, Y.~Liu, J.~Sun, F.~Jia, S.~Li, T.~Wang, and X.~Zhang, ``Cross modal transformer: Towards fast and robust 3d object detection,'' in \emph{Proceedings of the IEEE/CVF international conference on computer vision}, 2023, pp. 18\,268--18\,278.

\end{thebibliography}

% \begin{thebibliography}{00}
% \bibitem{b1} G. Eason, B. Noble, and I. N. Sneddon, ``On certain integrals of Lipschitz-Hankel type involving products of Bessel functions,'' Phil. Trans. Roy. Soc. London, vol. A247, pp. 529--551, April 1955.

% \end{thebibliography}

\vspace{12pt}

\end{document}